%% file: main.tex
\documentclass[sigconf, anonymous=false]{acmart}

\usepackage{booktabs} 
\usepackage[colorinlistoftodos]{todonotes}

\usepackage{csquotes}

\usepackage{epsfig}
\usepackage{color}

\usepackage{balance}  
\usepackage{algorithmic}
\usepackage{algorithm}
\usepackage{multirow}
\usepackage{graphicx}
\usepackage{amsmath}
\usepackage{amssymb}
\usepackage{amsfonts}
\usepackage{xspace}
\usepackage{url}

\newcommand{\IGNORE}[1]{}

\newcommand{\commentedtext}[1]{}

\def\CC{{C\nolinebreak[4]\hspace{-.05em}\raisebox{.4ex}{\tiny\bf ++}}}

\graphicspath{{./figures/}}

\newenvironment {squishlist}
{\begin{list}{$\bullet$}
		{ \setlength{\itemsep}{0pt}
			\setlength{\parsep}{3pt}
			\setlength{\topsep}{3pt}
			\setlength{\partopsep}{0pt}
			\setlength{\leftmargin}{1.5em}
			\setlength{\labelwidth}{1em}
			\setlength{\labelsep}{0.5em} } }
	{\end{list}}

\setcitestyle{numbers,sort&compress}

\newcommand{\example}{\ensuremath{\mathbf{x}}}
\newcommand{\bitstring}{\ensuremath{\mathbf{v}}}

\copyrightyear{2019}
\acmYear{2019}
\setcopyright{rightsretained}
\acmConference[SIGIR '19]{Proceedings of the 42nd International ACM SIGIR Conference on Research and Development in Information Retrieval}{July 21--25, 2019}{Paris, France}
\acmBooktitle{Proceedings of the 42nd International ACM SIGIR Conference on Research and Development in Information Retrieval (SIGIR '19), July 21--25, 2019, Paris, France}\acmDOI{10.1145/3331184.3331331}
\acmISBN{978-1-4503-6172-9/19/07}

\begin{document}

\title{Block-distributed Gradient Boosted Trees}


\author{Theodore Vasiloudis}
\authornote{Work performed during an internship at Amazon.}
\affiliation{
	\institution{RISE AI}
	\streetaddress{Stockholm, Sweden}}
\email{tvas@sics.se}

\author{Hyunsu Cho}
\affiliation{
	\institution{Amazon Web Services}
	\streetaddress{East Palo Alto, CA, USA}}
\email{chyunsu@amazon.com}

\author{Henrik Bostr\"{o}m}
\affiliation{
	\institution{KTH Royal Institute Of Technology}
	\streetaddress{Stockholm, Sweden}}
\email{bostromh@kth.se}

\renewcommand{\shortauthors}{T. Vasiloudis et al.}

\begin{abstract}
	\input{abstract.tex}
\end{abstract}

\maketitle

\section{Introduction}
\label{sec:introduction}
\input{introduction.tex}

\section{Background}
\label{sec:background}
\input{background.tex}

\section{Method}
\label{sec:method}
\input{method.tex}

\section{Evaluation}
\label{sec:evaluation}
\input{evaluation.tex}


\section{Conclusions}
\label{sec:conclusions}
\input{conclusions.tex}


%
\bibliographystyle{ACM-Reference-Format}
\bibliography{main}

\end{document}

%% file: abstract.tex
The Gradient Boosted Tree (GBT) algorithm is one of the most popular
machine learning algorithms used in production, for tasks that include
Click-Through Rate (CTR) prediction and learning-to-rank.
To deal with the massive datasets available today, many distributed GBT methods
have been proposed. However, they all
assume a row-distributed dataset, addressing scalability only with respect to the number of data points and not the number of features, and increasing
communication cost for high-dimensional data.

In order to allow for scalability across both the data point and feature
dimensions, and reduce communication cost, we propose block-distributed GBTs.
We achieve communication efficiency by making full use of the data sparsity and adapting the Quickscorer algorithm to the block-distributed setting.
We evaluate our approach using datasets with millions of features,
and demonstrate that we are able to achieve multiple orders of magnitude reduction
in communication cost for sparse data, with no loss in accuracy, while providing a more scalable
design.
As a result, we are able to reduce the training time for high-dimensional data,
and allow more cost-effective scale-out without the need for expensive network
communication.

%% file: introduction.tex
In some of the most important Information Retrieval (IR) tasks like CTR prediction and learning-to-rank,
Gradient Boosted Trees have risen as one of the most effective algorithms,
due to their accuracy and scalability \cite{mcrank, ctr-facebook}.


For massive amounts of data that are common at web-scale,
distributed learning methods have become necessary.
In such scenarios, we want to have the best possible utilization of resources and scalability, to minimize the cost of training a new model.


Currently, distributed GBT algorithms \cite{xgboost, lightgbm, dimboost} use row distribution,
in which every worker node processes a horizontal slice of the data matrix, i.e. a subset of the data points with all feature values.
Row distribution enables scalability only with respect to the number of data points. As a result, for high-dimensional data with millions of
features, existing algorithms require large amounts of memory, which we can only remedy
by adding worker nodes to the cluster, further increasing the amount of network communication.

In addition, the existing algorithms use dense communication to aggregate gradient histograms
between the workers.
Algorithms like XGBoost~\cite{xgboost} and LightGBM~\cite{lightgbm}
use all-reduce~\cite{allreduce} or similar approaches, where
we would need to communicate millions of histograms, one per feature, regardless
of whether a feature is present in a data partition or not.
As a result,
the communication cost may become impractical for sparse, high-dimensional data. The issue is more pressing for
multi-tenant clusters or cloud environments commonly used for training \cite{comm-efficient-gbt},
since the GBT training job needs to compete with other running jobs for network bandwidth.


To address these issues, we introduce a block-distributed GBT algorithm that partitions the training data
in both the row and column dimensions.
The new algorithm adds a second
dimension along which we can scale-out our learning and uses a sparse
gradient histogram representation to minimize network communication.
Unlike existing algorithms for feature-parallel training \cite{lightgbm, xgboost}, our proposed
algorithm does not assume that the complete dataset would fit into the main memory of each
worker node.


Block distribution introduces a new algorithmic challenge: the worker nodes no longer
have complete access to all feature values.
The regular method of computing prediction outputs for decision trees, top-to-bottom traversal,
no longer works, because only a subset of feature values are visible to each worker. We
develop a novel use of the Quickscorer algorithm \cite{quickscorer} for block-distributed prediction,
to make prediction using parts of each data point feasible and efficient.

However, the new method for calculating tree outputs
introduces additional communication steps in the training pipeline.
Careful design is therefore necessary to offset the cost.
To that end, our proposed algorithm uses the parameter server \cite{muPS} to efficiently transmit sparse data, unlike the dense communication used by current methods~\cite{dimboost, xgboost, lightgbm}.
Although the use of the parameter server adds overhead for dense, low-dimensional data, compared to all-reduce, it results in orders of magnitude reduced
communication cost for sparse, high-dimensional data.

To summarize, our contributions are the following:

\begin{squishlist}

	\item We propose the first block-distributed GBT algorithm, allowing for scalability
	across both the data point and feature dimensions.
	\item  We provide an experimental evaluation on datasets with up to millions of
	features. We illustrate the advantages and limitations of the approach in sparse
	and dense settings.
	\item  We demonstrate orders of magnitude improved communication cost by taking full
	advantage of data sparsity during training.
\end{squishlist}

%% file: background.tex
In this section we will provide a brief introduction to gradient
boosting and the relevant parts of row-distributed Gradient Boosted
Tree learning. We refer the reader to \cite{boosting-survey} for
an in-depth survey of gradient boosting.

%
%

\subsection{Gradient Boosted Trees}

GBT learning algorithms all follow a similar base algorithm.
At each iteration, we first make predictions on the training data using the
current ensemble. We then get the gradients for each data point according to our loss function, and use those gradients to determine the \emph{gradient histograms}
for every feature, at every leaf.
Finally, we grow the tree by finding for each leaf its optimal split
point.

The gradient histograms for a leaf-feature combination, determine what
would be the gradient distribution for the left and right child of that
leaf, if we were to split the leaf using that feature, for every possible
value of the feature. Our aim as we optimize the tree is to select splits
that maximize the gain made possible by splitting a leaf, in terms of a loss function,
like squared loss \cite{boosting-survey}.

Doing this for every possible feature and value quickly becomes untenable for
real-valued features, so most algorithms make use of \emph{quantized histograms} \cite{xgboost}.
We determine a number of buckets $B$, create empirical histograms for each feature to determine
candidate split points, and iterate through
this pre-determined number of split points for every feature when evaluating splits.
We can then select the optimal split for each leaf by simply sorting the
feature-value combinations.


\subsection{Parameter Server \& All-Reduce}

In distributed machine learning there are two dominant communication patterns,
each with its merits and drawbacks. The all-reduce primitive performs an aggregation
across the cluster, at the end of which, every worker in the cluster ends up with the
same copy of the aggregated data. Algorithms like a binomial tree all-reduce \cite{allreduce} are used to minimize communication.
However, all-reduce uses dense communication and requires the byte size of the parameters
being communicated to be known in advance, resulting in redundancies in
the case of sparse data.

The parameter server architecture \cite{muPS} on the other hand has machines play
different roles. \emph{Workers} are responsible for computation and communicate with \emph{servers}
responsible for storing and updating parameters. The parameter server is more flexible than
an all-reduce approach, at the cost of added complexity.

\subsection{Row-distributed Gradient Boosted Trees}
\label{sec:background-row-distributed}

In the row-distributed setting each worker gets a \emph{horizontal partition} of the
data, i.e. a subset of the data points with all features included, makes predictions, and calculates the
local gradient histograms for its partition. These gradients need then to
be aggregated with every other worker so that they all end up with the same
combined gradient histograms. Once the gradients have been communicated, each
worker can use its local copy to find the optimal splits, and update the model.

Row-distributed methods utilize dense communication:
The algorithms will, for every leaf and every feature, communicate $B$ values, where $B$ the maximum bucket count, regardless of the actual number of unique values
for that leaf-feature combination.

For example, assume a continuous feature that we quantize to 255 buckets,
which is the default choice for XGBoost. As we grow the tree and partition
the dataset into smaller groups, it becomes more likely that many of these
255 value ranges will not be represented in the partition present in
the leaf. However, the all-reduce
methods require the number of values to be communicated to be known in advance, so we will end up communicating for every leaf the full
255 values, despite the fact that many of those values will be zero.
This problem is further magnified in sparse datasets that are common
in CTR and learning-to-rank tasks.

Through our use of sparse histogram representations we aim to overcome this shortcoming,
by communicating for each leaf only the strictly necessary values in a histogram,
vastly improving the communication cost for sparse datasets.

%% file: method.tex
In this section we provide an overview of the methods we developed
to make block-distributed GBT training possible, and the optimizations
that make the process efficient.
We use the parameter server for network communication. In this setting, the worker
nodes have access to the data and perform computations, like calculating the
local gradient histograms, while the servers are responsible for aggregating
the parameters sent by the workers.

\subsection{Block-distributed prediction}
\label{sec:method-prediction}

In the block-distributed setting every worker has access to a block
of the data matrix, that is, data are sliced both horizontally and
vertically, and one data block is assigned to each worker.

In order to determine the exit leaf $e$, that is, the leaf where a data point ends up in
when dropped down a tree,
we will need to communicate between the workers that share different parts of the same rows
in the dataset. To minimize the communication cost and achieve provably correct
predictions we make use of the Quickscorer (QS) \cite{quickscorer} algorithm that we adapt to
work in the block-distributed setting. We briefly describe the algorithm here and
refer the reader to the original text
for an in-depth description.

\subsubsection*{Quickscorer overview}

The QS algorithm works by assigning a bitstring to every node in a tree,
that indicates which leafs would be removed from the set of candidate exit leafs
whenever the condition in the node is evaluated as false.
Every internal node in the tree is assigned a bitstring of length $\vert L \vert$,
where $L$ is the set of leaves in the tree. The bitstring of a node has zeros
at the bits corresponding to the leaves that would become impossible to reach if
the node's condition evaluates to false, and ones everywhere else.

We assume that
there exists an oracle \texttt{FindFalse} that, given an input $\example$, returns the set of
all internal nodes whose conditions
evaluate to false for $\example$, without the need to evaluate all the associated test
conditions. Once we have that set, we initialize bit vector $\bitstring$ with all 1's
and then update $\bitstring$ by performing bitwise-\texttt{AND}
with every bitstring in the set. \citet{quickscorer} prove
that the exit leaf $e$ corresponds to the left-most bit set to 1 in the updated $\bitstring$.

\subsubsection*{Block-distributed Quickscorer}

We adapt the above algorithm to work in the case where each worker only has
access to a subset of the features for each data point. To fully determine the exit node $e$ for a
single data point $\example$, we need to combine the predictions from multiple workers,
that hold the different parts of the data point.
Since each worker only has a range of features available, it can only evaluate
the conditions for the features it has available.
We use Quickscorer to only communicate prediction bitstrings, achieving significant communication savings
compared to communicating complete models or data points.

Let $N$ be the set of internal nodes in a tree. Each internal node contains
a condition of the form $\example[f_n] \leq \gamma$, where $f_n$ the
selected feature for the node and $\gamma$ the threshold.
Let $W$ be the set of workers, and $S$ the set of servers.
Each worker $w$ will only be able to
evaluate the condition for a subset of the nodes $N_w$, where $\cup_w{N_w} = N$.

For example, say we have ten features in the data, and three of those,
$\{f_1, f_2, f_8\}$, are conditions in the tree. Let worker 1 be responsible
for features 1--5 and worker 2 responsible for features 6--10. Then to fully
determine the exit leaf for a data point $\example$ we would evaluate conditions $\{f_1, f_2\}$ on worker 1 and condition $\{f_8\}$ on worker 2. Each
worker then pushes their partial bitstrings $\{\bitstring_1, \bitstring_2\}$ to a server, which performs
the bit-wise \texttt{AND} operation $\bitstring = \bitstring_1 \land \bitstring_2$ that determines the exit leaf for
$\example$.

Due to the commutative
nature of the \texttt{AND} operator, the order in which these aggregations happen does not
matter. Because \texttt{AND} is also associative, the overall conjunction of the
partial $\bitstring_{w}$'s will be equivalent to the overall $\bitstring$. As a result of
these properties, the block-distributed Quickscorer will have provably correct predictions.

In our implementation, each server is responsible for a horizontal slice of the data.
When a worker pushes a vector of bistrings to a server, the server
performs a bitwise \texttt{AND} for each data point. Once all workers
that belong to a horizontal data slice have pushed their bitstrings,
we can use the aggregated bitstrings on the servers to determine the exit leaves.
The workers can finally pull the exit node ids, make the predictions, and prepare to calculate the
gradients for the next step.

\subsection{Block-distributed histogram aggregation}
\label{sec:method-aggregation}


In the block distributed setting each worker will only have a partial view of each data point,
so it can only calculate gradients histograms for the data points and range of features it has available.
However, since
we are performing a simple sum operation over all the gradients, which is a commutative and
associative operation, we can perform the local sums at each worker first, push the partial
gradients to the servers, and perform the feature-range sum at each server to get the complete
histograms for \emph{each feature range}.

Our communication pattern now changes from the prediction step:
each server is now responsible and aggregates the statistics for a range of \emph{features} instead of \emph{data points}.

We can think of the gradient histograms as a \textit{sparse} tensor with dimensions $|L| \times |F| \times
B$, where $L$ the set of leaves, $F$ the set of features, and $B$ the number of possible split
points. Each worker is responsible for populating a block of this tensor.

For every data point in the block we use the exit nodes from the previous step to get a prediction and gradient value. Given this leaf-to-example mapping, we iterate through each leaf, each
data point that belongs to that leaf, and through each feature that belongs to this block. Given
the feature index and value, we find the corresponding bucket in the gradient histogram and add the
data point's gradient to that bucket, indexed as $(i,j,k)$ where $i$ is the leaf id, $j$ is the feature
id, and $k$ is the histogram bucket that corresponds to the feature value.

Once each worker has populated their block of the tensor, they will push their partial tensor to a
specific server. Each server is responsible for a range of features, so all workers
that have access to the same \emph{vertical} slice of data will send their data to the same server.

Note that each worker will communicate \emph{at most} $|L| \times (|F| / |S|) \times B$ elements of the sparse tensor, where $|S|$ the number of servers. However, because we are using a sparse
representation, none of the zero values in the histograms will be communicated, making this step
potentially much more efficient than existing approaches that always communicate $|L| \times |F| \times B$ elements per worker.

On the server side, every time a new sparse tensor arrives it is simply summed together with the
existing one, where the initial element is a tensor of all zeros. Once all workers
for a vertical slice have sent their partial sums to their corresponding server, that server will
have a complete view of the gradient histograms for the range of features it is responsible for,
and we can proceed to the final step of split finding.
For this, we use an algorithm similar to the
one described in \cite{dimboost}, so we omit its description due to space limitations.

%% file: evaluation.tex
In this section we present an evaluation of our block-distributed approach in
terms of communication efficiency and how that translates to end-to-end runtime.

\subsection{Experimental Setup}

To ensure a fair comparison between the methods we implement both the
block-distributed and row-distributed algorithms in \CC. We make use
to the parameter server originally developed by \citeauthor{muPS} for the
block-distributed implementation
and use the Rabit \cite{rabit} all-reduce framework for the row-distributed version.

We use a local cluster of 12 workers. We use all twelve as workers for the row-distributed
experiments, and use 9 as workers and 3 as servers for the block-distributed experiments.

We use four large-scale binary classification datasets
\footnote{Data available at \url{https://www.csie.ntu.edu.tw/~cjlin/libsvmtools/datasets/binary.html}}
with different sparsity characteristics to evaluate
the performance of each approach.
The first two, URL and avazu, are extremely sparse with approximately 3.2 million and 1 million features
respectively. RCV1 has approximately 47 thousand features, and Bosch is a dense dataset with 968
features. Since we focus on feature density we train on a 20,000 data point sample from each
dataset.

Our main metric is communication cost, measured in the amount of data the being communicated
for the gradient histograms, in MiB. In addition, to measure the real-world performance of the
approach we compare the end-to-end runtime as the combined computation and communication time
of the histogram aggregation step.

\subsection{Results}

\begin{figure}
	\includegraphics[width=\columnwidth]{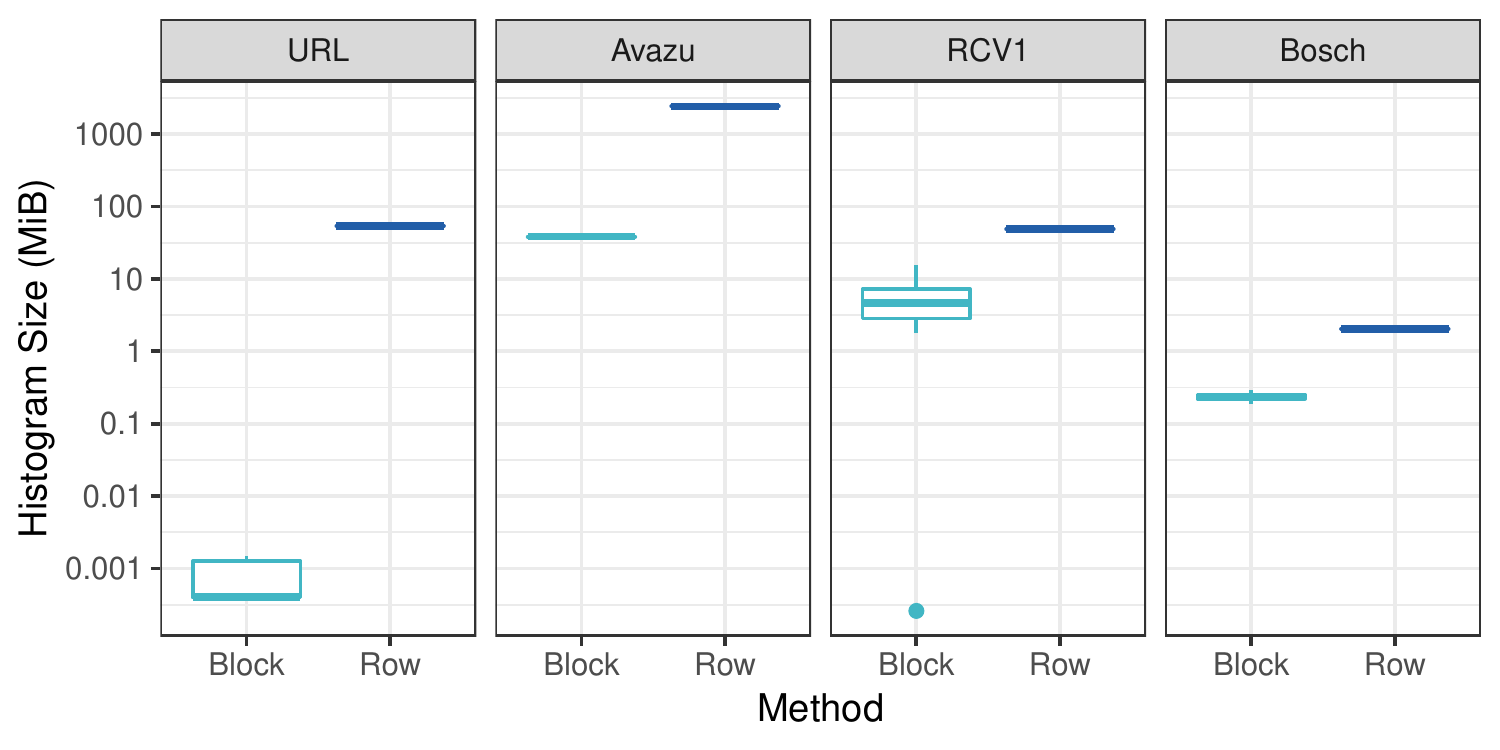}
  \vspace{-20pt}
	\Description{A graph showing that the amount of data communicated for the datasets
		are from one to five orders of magnitude smaller for the method described in this
		paper, compared to existing row-distributed approaches.}
	\caption{The byte size of the gradient histograms being communicated for the various datasets.}
	\label{fig:hist-size}
\end{figure}

We start with communication cost, illustrated in
Figure \ref{fig:hist-size}. From the figure, where the y-axis is in log scale, we can see the significant communication
savings that the sparse histogram representation brings: the histograms produced by the
row-distributed algorithms are one to five orders of magnitude larger than the sparse ones developed
for this study.

\begin{figure}
	\includegraphics[width=\columnwidth]{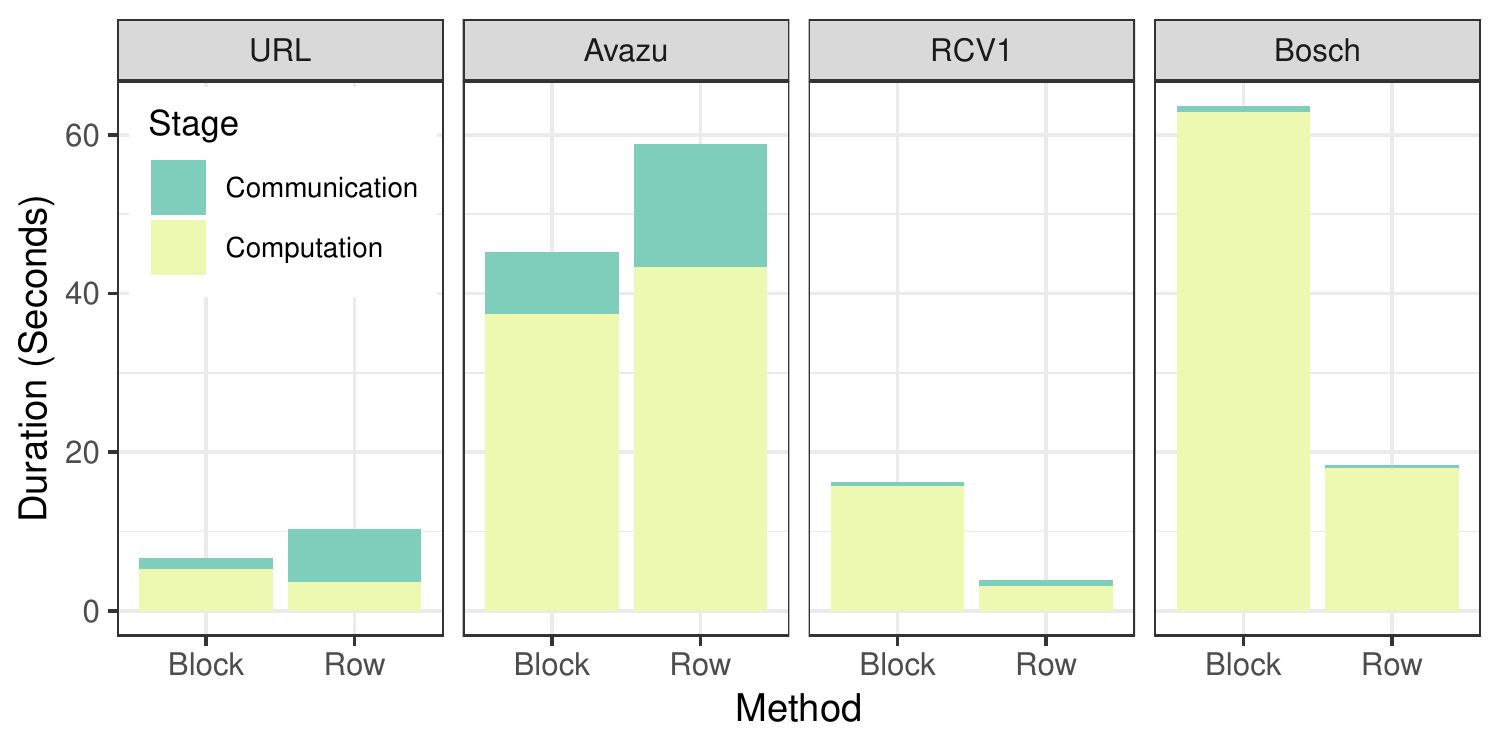}
  \vspace{-20pt}
	\Description{A graph showing that the total runtime is smaller for the sparse datasets
		for the methods in this paper, but larger for the less sparse or dense dataset.}
	\caption{The communication and computation times for the various datasets, in seconds.}
	\label{fig:time}
  \vspace{-10pt}
\end{figure}

The communication time is therefore significantly reduced for the block-distributed approach, however
there is an increase in computation time. This is caused by the overhead introduced
by the use of the parameter server and sparse tensors. Unlike dense data structures that are contiguous
in memory, the sparse tensors require indirect addressing, resulting in an implementation that is not
cache-friendly. In addition, the parameter server approach requires us to pack and unpack
the sparse tensors into contiguous arrays of floating point numbers, which are then serialized
to be sent over the network and de-serialized server-side. This all creates a significant overhead, as
evidenced by our experiments on the less sparse RCV1 and Bosch datasets where the computation phase dominates the communication time, as shown in Figure \ref{fig:time}.

%% file: conclusions.tex
In this study we presented the first block-distributed algorithm for
Gradient Boosted Tree training. Our design allows us to scale-out learning across
both the data point and feature dimensions, allowing for cheaper hardware
to be used to train massive, sparse datasets. Through our use of a sparse
representation of the gradient histograms, we are able to reduce the
communication cost of the algorithm by multiple orders of magnitude for sparse
data.
However, the approach introduces computational overhead that makes it inefficient
for datasets with limited sparsity.

In future work we aim to overcome this limitation by using more cache-friendly sparse tensor representations.
One benefit of the parameter server is the
ability to overlap communication with computation, e.g. we could be performing
the split finding server side while the gradients are being sent by the workers.
Finally, we can make use of the information from previous iterations to prioritize
features that more likely to be the best split point, and make use of the Hoeffding
bound to stop split finding early.